\documentclass{article}
\usepackage{iclr2021_conference,times}

\usepackage{hyperref}
\definecolor{mydarkblue}{rgb}{0,0.08,0.45}
\hypersetup{colorlinks=true, linkcolor=mydarkblue, citecolor=mydarkblue, filecolor=mydarkblue, urlcolor=mydarkblue, pdfview=FitH}
\usepackage{url}

\usepackage{float}
\usepackage[utf8]{inputenc} 
\usepackage[T1]{fontenc}    
\usepackage{hyperref}       
\usepackage{url}            
\usepackage{booktabs}       
\usepackage{amsfonts}       
\usepackage{nicefrac}       
\usepackage{microtype}      
\usepackage{multirow}       
\usepackage{amssymb}        
\usepackage{spverbatim}     
\usepackage{wrapfig}
\usepackage{pifont}
\usepackage{amsmath}
\usepackage[shortlabels]{enumitem}
\usepackage{caption}
\usepackage{wrapfig}
\usepackage{subcaption}

\usepackage{graphicx,tikz,pgfplots,pgfplotstable,pgfpages}
\usepgfplotslibrary{groupplots}
\usepgfplotslibrary{fillbetween}
\definecolor{C0}{RGB}{31, 119, 180}
\definecolor{C1}{RGB}{255, 127, 14}
\definecolor{C2}{RGB}{44, 160, 44}
\definecolor{C3}{RGB}{214, 39, 40}
\definecolor{C4}{RGB}{148, 103, 189}
\definecolor{C5}{RGB}{140, 86, 75}
\definecolor{C6}{RGB}{227, 119, 194}
\definecolor{C7}{RGB}{127, 127, 127}
\definecolor{C8}{RGB}{188, 189, 34}
\definecolor{C9}{RGB}{23, 190, 207}
\pgfplotscreateplotcyclelist{default}{C0, C1, C2, C3, C4, C5, C6, C7, C8, C9}
\definecolor{kngreen}{RGB}{106,148,81}
\definecolor{knblue}{RGB}{55,88,136}
\definecolor{knyellow}{RGB}{222,162,81}
\definecolor{knred}{RGB}{175,57,55}
\definecolor{knviolet}{RGB}{104,65,118}
\definecolor{kngrey}{RGB}{125,125,125}

\usepackage[eulergreek]{sansmath}
\usepgfplotslibrary{colorbrewer}
\pgfplotsset{compat=1.16}
\pgfplotsset{
    ticklabel style={font=\scriptsize\sffamily\sansmath},
        every axis label/.append style={font=\scriptsize\sffamily\sansmath},
        legend style={font=\scriptsize\sffamily\sansmath, /tikz/every even column/.append style={column sep=0.15cm}},
        title style={font=\footnotesize\sansmath, yshift=-3pt},
        cycle list name=default,
}

\title{Graph-Based Continual Learning}

\author{Binh Tang \\
    Department of Statistics and Data Science\\
    Cornell University\\
    Ithaca, NY 14850 \\
    \texttt{bvt5@cornell.edu} \\
    \And
    David S. Matteson \\
    Department of Statistics and Data Science \\
    Cornell University \\
    Ithaca, NY 14850 \\
    \texttt{matteson@cornell.edu} \\
}

\iclrfinalcopy
\begin{document}

\maketitle

\begin{abstract}
	Despite significant advances, continual learning models still suffer from catastrophic forgetting when exposed to incrementally available data from non-stationary distributions. Rehearsal approaches alleviate the problem by maintaining and replaying a small episodic memory of previous samples, often implemented as an array of independent memory slots. In this work, we propose to augment such an array with a learnable random graph that captures pairwise similarities between its samples, and use it not only to learn new tasks but also to guard against forgetting. Empirical results on several benchmark datasets show that our model consistently outperforms recently proposed baselines for task-free continual learning.
\end{abstract}

\section{Introduction}

Recent breakthroughs of deep neural networks often hinge on the ability to repeatedly iterate over stationary batches of training data. When exposed to incrementally available data from non-stationary distributions, such networks often fail to learn new information without forgetting much of its previously acquired knowledge, a phenomenon often known as \textit{catastrophic forgetting} \citep{ratcliff1990connectionist,mccloskey1989catastrophic,french1999catastrophic}. Despite significant advances, the limitation has remained a long-standing challenge for computational systems that aim to continually learn from dynamic data distributions \citep{parisi2019continual}.

Among various proposed solutions, rehearsal approaches that store samples from previous tasks in an episodic memory and regularly replay them are one of the earliest and most successful strategies against catastrophic forgetting \citep{lin1992self,rolnick2019experience}. An episodic memory is typically implemented as an array of independent slots; each slot holds one example coupled with its label. During training, these samples are interleaved with those from the new task, allowing for simultaneous multi-task learning as if the resulting data were independently and identically distributed.

While such approaches are effective in simple settings, they require sizable memory and are often impaired by memory constraints, performing rather poorly on complex datasets. A possible explanation is that slot-based memories fail to utilize relational structure between samples; semantically similar items are treated independently both during training and at test time. In marked contrast, relational memory is a prominent feature of biological systems that has been strongly linked to successful memory retrieval and generalization \citep{prince2005neural}. Humans, for example, encode event features into cortical representations and bind them together in the medial temporal lobe, resulting in a durable, yet flexible form of memory \citep{shimamura2011episodic}.

In this paper, we introduce a novel Graph-based Continual Learning model (GCL) that resembles some characteristics of relational memory. More specifically, we explicitly model pairwise similarities between samples, including both those in the episodic memory and those found in the current task. These similarities allow for representation transfer between samples and provide a resilient mean to guard against catastrophic forgetting. Our contributions are twofold:

\begin{enumerate}[(1)]
	\item We propose the use of random graphs to represent relational structures between samples. While similar notions of dependencies have been proposed in the literature \citep{louizos2019functional,yao2020automated}, the application of random graphs in task-free continual learning is novel, at least to the best of our knowledge.
	\item We introduce a new regularization objective that leverages such random graphs to alleviate catastrophic forgetting. In contrast to previous work \citep{rebuffi2017icarl,li2017learning} based on knowledge distillation \citep{hinton2015distilling}, the objective penalizes the model for forgetting learned edges between samples rather than their output predictions.
\end{enumerate}

Our approach performs competitively on four commonly used datasets, improving accuracy by up to 19.7\% and reducing forgetting by almost 37\% in the best case when bench-marked against competitive baselines in task-free continual learning.

\begin{figure}[t]
	\centering
	\includegraphics[trim=3.5cm 24cm 13cm 0cm,clip,width=\linewidth]{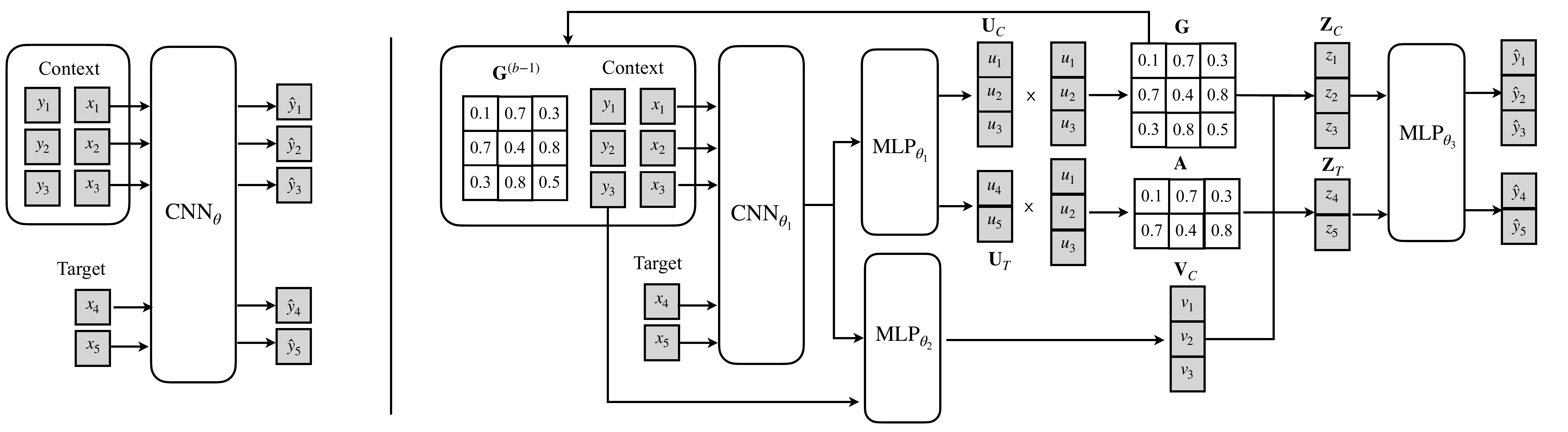}
	\caption{\small Illustration of Experiment Replay (ER) \citep{chaudhry2019continual} on the \textit{left} and our model (GCL) on the \textit{right}. While ER independently processes \textit{context images} from the episodic memory and \textit{target images} from the current task, GCL models pairwise similarities between the images via the random graphs $\mathbf{G}$ and $\mathbf{A}$.}
	\label{fig:diagram}
\end{figure}

\section{Problem Formulation}
In this work, we follow the learning protocol for image classification from \cite{lopez2017gradient}. More specifically, we consider a training set $\mathcal{D} = \{ \mathcal{D}_1, \cdots, \mathcal{D}_{T} \}$ consisting of $T$ tasks where the dataset for the $t$-th task $\mathcal{D}_{t} = \{(\mathbf{x}_i^t, \mathbf{y}_i^t)\}_{i=1}^{n_t}$ contains $n_t$ input-target pairs $(\mathbf{x}_i^t, \mathbf{y}_i^t) \in \mathcal{X} \times \mathcal{Y}$. While the tasks arrive sequentially and exclusively, we assume the input-target pairs $(\mathbf{x}_i^t, \mathbf{y}_i^t)$ in each task are independent and identically distributed (i.i.d.). The goal is to learn a supervised model $f_\theta: \mathcal{X} \rightarrow \mathcal{Y}$, parametrized by $\theta$, that outputs a class label $\mathbf{y} \in \mathcal{Y}$ given an unseen image $\mathbf{x} \in \mathcal{X}$.

Following prior work \citep{lopez2017gradient,riemer2018learning,chaudhry2019continual}, we consider online streams of tasks in which samples from different tasks arrive at different times. As an additional constraint, we insist that the model can only revisit a small amount of data chosen to be stored in a fixed-size episodic memory $\mathcal{M}$.

For clarity, we refer to the data in such an episodic memory as \textit{context images} and \textit{context labels} and denote by $\mathbf{X}_{\mathcal{C}} = \{\mathbf{x}_i\}_{i \in \mathcal{C}}$ and $\mathbf{Y}_{\mathcal{C}} = \{\mathbf{y}_i\}_{i \in \mathcal{C}}$, respectively.
These images and labels are to be distinguished from those in the current task, which we refer to as \textit{target images} and \textit{target labels} and denote by $\mathbf{X}_{\mathcal{T}} = \{\mathbf{x}_j\}_{j \in \mathcal{T}}$ and $\mathbf{Y}_{\mathcal{T}} = \{\mathbf{y}_j\}_{j \in \mathcal{T}}$, respectively. While the model is allowed to update the context samples during training, the episodic memory is necessarily frozen at test time.

\section{Graph-Based Continual Learning}

In this section, we propose a Graph-based Continual Learning (GCL) algorithm. While most rehearsal approaches ignore the correlations between images and independently pass them through a network to compute predictions \citep{rebuffi2017icarl,chaudhry2019continual,aljundi2019gradient}, we model pairwise similarities between the images with learnable edges in random graphs (see Figure \ref{fig:diagram}). Intuitively, although it might be easy for the model to forget any particular sample, the multiple connections it forms with similar neighbors are harder to be forgotten altogether. If trained well, the random graphs can therefore equip the model with a plastic and durable means to fight against catastrophic forgetting.

\smallskip
\textbf{Graph Construction.} \enskip Given a minibatch of target images $\mathbf{X}_{\mathcal{T}}$ from the current task, our model makes predictions based on the context images $\mathbf{X}_{\mathcal{C}}$ and context labels $\mathbf{Y}_{\mathcal{C}}$ that span several previously seen tasks, up to and including the current one. In particular, we explicitly build two random graphs of pairwise dependencies: an undirected graph $\mathbf{G}$ between the context images $\mathbf{X}_{\mathcal{C}}$ and a directed, bipartite graph $\mathbf{A}$ from the context images $\mathbf{X}_{\mathcal{C}}$ to the target images $\mathbf{X}_{\mathcal{T}}$.

Since an undirected graph can be thought of as a directed graph between its vertices and a copy of itself, we treat the \textit{context graph} $\mathbf{G}$ as such and build it analogously to the \textit{context-target graph} $\mathbf{A}$. Specifically, the high-dimensional context images $\mathbf{X}_{\mathcal{C}}$ and target images $\mathbf{X}_{\mathcal{T}}$ are first mapped to the image embeddings $\mathbf{U}_{\mathcal{C}}$ and $\mathbf{U}_{\mathcal{T}}$, respectively, using an image encoder $f_{\theta_1}: \mathcal{X} \rightarrow \smash{\mathbb{R}^{d_1}}$. Following \cite{louizos2019functional}, we then represent the edges in each graph by independent Bernoulli random variables whose means are specified by a kernel function in the embedding space. More precisely, the distribution of the resulting Erd\H{o}s-R\'enyi random graphs \citep{erdos59a} can be defined as
\begin{align}
	p(\mathbf{G} \, \lvert \, \mathbf{U}_{\mathcal{C}})                           & = \prod_{i \in \mathcal{C}} \prod_{k \in \mathcal{C}} \text{Ber}(\mathbf{G}_{ik} \, \lvert \, \kappa_\tau(\mathbf{u}_i, \mathbf{u}_k)), \label{pg} \\
	p(\mathbf{A} \, \lvert \, \mathbf{U}_{\mathcal{T}}, \mathbf{U}_{\mathcal{C}}) & = \prod_{j \in T} \prod_{k \in \mathcal{C}} \text{Ber}(\mathbf{A}_{jk} \, \lvert \, \kappa_\tau(\mathbf{u}_j, \mathbf{u}_k)), \label{pa}
\end{align}
for all $i, k \in \mathcal{C}$ and $j \in \mathcal{T}$ where $\kappa_\tau: \mathbb{R}^{d_1} \times \mathbb{R}^{d_1} \rightarrow [0, \infty)$ is a kernel function that encodes similarities between image embeddings such as the RBF kernel $\kappa_\tau(\mathbf{u}_i, \mathbf{u}_j) = \smash{\exp\left(-\frac{\tau}{2}\|\mathbf{u}_i - \mathbf{u}_j\|_2^2  \right)}$. Here, with a slight abuse of notation, we also use $\mathbf{G}$ and $\mathbf{A}$ to denote the corresponding adjacency matrices; $\textbf{A}_{jk} \in \{0, 1\}$, for example, represents the presence or absence of a directed edge between the $j$-th target image and the $k$-th context image.

\smallskip
\textbf{Predictive Distribution.} \enskip Given a context graph $\mathbf{G}$ and a context-target graph $\mathbf{A}$ that encode pairwise similarities to the context images, our next step is to propagate information from the context images $\mathbf{X}_{\mathcal{C}}$ and context labels $\mathbf{Y}_{\mathcal{C}}$ to make predictions. To that end, we embed $\mathbf{X}_{\mathcal{C}}$ by another image encoder $f_{\theta_2}$ with weights partially tied to the previous one $f_{\theta_1}$, and encode $\mathbf{Y}_{\mathcal{C}}$ by a linear label encoder before concatenating the resulting embeddings into latent representations $\mathbf{V}_\mathcal{C} \in \smash{\mathbb{R}^{|\mathcal{C}| \times d_2}}$. In combination with the distributions of $\mathbf{G}$ and $\mathbf{A}$, we compute context-aware representations for the context images and target images, denoted by $\{\mathbf{z}_i\}_{i \in \mathcal{C}}$ and $\{\mathbf{z}_j\}_{j \in \mathcal{T}}$, respectively:
\begin{align}
	p(\mathbf{z}_i \, \lvert \, \mathbf{U}_{\mathcal{C}}, \mathbf{V}_{\mathcal{C}})                           & = \int_{\mathbf{G}} \mathbb{I}_{\{\tilde{\mathbf{G}}_i \mathbf{V}_{\mathcal{C}}\}} (\mathbf{z}_i) \, dP(\mathbf{G} \, \lvert \, \mathbf{U}_{\mathcal{C}}) \label{pzc}                            \\
	p(\mathbf{z}_j \, \lvert \, \mathbf{U}_{\mathcal{T}}, \mathbf{U}_{\mathcal{C}}, \mathbf{V}_{\mathcal{C}}) & = \int_{\mathbf{A}} \mathbb{I}_{\{\tilde{\mathbf{A}}_j \mathbf{V}_{\mathcal{C}}\}} (\mathbf{z}_j) \, dP(\mathbf{A} \, \lvert \, \mathbf{U}_{\mathcal{T}}, \mathbf{U}_{\mathcal{C}}). \label{pzt}
\end{align}
where $\tilde{\mathbf{G}}_i$ and $\tilde{\mathbf{A}}_j$ indicate the $i$-th and $j$-th row of $\mathbf{G}$ and $\mathbf{A}$, each normalized to sum to 1, and $\mathbb{I}_{\mathcal{S}}(\cdot)$ denotes the indicator function on a set $\mathcal{S}$. Intuitively, the representations $\mathbf{V}_\mathcal{C}$ are linearly weighted by each graph sample, and the normalization step ensures proper scaling in case the numbers of edges formed with the context images vary. Once we summarize each image by the context samples, a final network $f_{\theta_3}: \mathbb{R}^{d_2} \rightarrow \mathcal{Y}$ takes as input the context-aware representations and produces predictive distributions:
\begin{align}
	p(\mathbf{y}_i \, \lvert \, \mathbf{X}_{\mathcal{C}})               & = \int_{\mathbf{z}_i} p\left(\mathbf{y}_i \, \lvert \, f_{\theta_3}(\mathbf{z}_i)\right)  \, dP(\mathbf{z}_i \, \lvert \, \mathbf{U}_{\mathcal{C}}, \mathbf{V}_{\mathcal{C}}), \label{pyc}                         \\
	p(\mathbf{y}_j \, \lvert \, \mathbf{x}_j, \mathbf{X}_{\mathcal{C}}) & = \int_{\mathbf{z}_j} p\left(\mathbf{y}_j \, \lvert \, f_{\theta_3}(\mathbf{z}_j)\right) \, dP(\mathbf{z}_j\, \lvert \, \mathbf{U}_{\mathcal{T}}, \mathbf{U}_{\mathcal{C}}, \mathbf{V}_{\mathcal{C}}). \label{pyt}
\end{align}
Since the numbers of random binary graphs $\mathbf{G}$ and $\mathbf{A}$ are exponential, we approximate the integrals in (1) - (6) by Monte Carlo samples. More specifically, we use one sample of $\mathbf{G}$ and $\mathbf{A}$ during training and 30 samples of $\mathbf{A}$ during testing. Also, these graph samples are inherently non-differentiable, so we use the Gumbel-Softmax relaxations of the Bernoulli random variables during training \citep{maddison2016concrete,jang2016categorical}. The degree of approximation is controlled by temperature hyper-parameters, which exert significant influence over the density of the graph samples. We find that a small temperature for $\mathbf{G}$ and a larger temperature for $\mathbf{A}$ work well.

There are several reasons for making the graphs $\mathbf{G}$ and $\mathbf{A}$ random. First, the stochasticity induced by the Bernoulli random variables allows us to output multiple predictions and average these predictions, and such ensemble techniques have been quite successful in continual learning settings \citep{coop2013ensemble,fernando2017pathnet}. Perhaps more importantly, we find that the deterministic version with the Bernoulli random variables replaced by their parameters results in very sparse graphs where samples from the same classes are often deemed dissimilar. In a similar fashion to dropout \citep{srivastava2014dropout}, the random edges encourage the model to be less reliant on a few particular edges and therefore promote knowledge transfer between samples. By a similar reasoning, we remove self-edges in the context graph and also observe more connections between samples.

\textbf{Graph Regularization.} \enskip As training switches to new tasks, the distributional shifts to the target images necessarily result in changes to both the context graph $\mathbf{G}$ and the context-target graph $\mathbf{A}$. In addition, the context images are regularly updated to be representative of the data distribution up to that point, so any well-learned connections between the context images are also susceptible to catastrophic forgetting. As a remedy, we save the parameters of the Bernoulli edges to the episodic memory in conjunction with the context images and context labels, and introduce a regularization term that discourages the model from forgetting previously learned edges:
\begin{equation}
	\mathcal{L}^{(b)}_{\mathbf{G}}(\theta_1) \triangleq \frac{1}{|\mathcal{I}^{(b)}|} \, \ell\!\left(p\!\left(\mathbf{G}^{(b - 1)}_{\mathcal{I}^{(b)}}\right), \, p\!\left(\mathbf{G}^{(b)}_{\mathcal{I}^{(b)}}\right)\right). \label{eq:graph-regularization}
\end{equation}
Here, $\ell(\cdot, \cdot)$ denotes the cross-entropy between two probability distributions, $\smash{\mathcal{I}^{(b)}}$ the index set of edges to be regularized in the $b$th minibatch, and $\smash{\mathbf{G}^{(b - 1)}}$ the adjacency matrix learned from the beginning up to the previous minibatch. The selection strategies $\smash{\mathcal{I}^{(b)}}$ are discussed in the next subsection. Besides the regularization term, our training objective includes two other cross-entropy losses, one for the context images and another for the target images:
\begin{equation}
	\mathcal{L}(\theta_1, \theta_2, \theta_3) = \frac{\lambda_{\mathcal{C}}}{|\mathcal{C}|} \sum_{i \in \mathcal{C}} \ell\!\left(\mathbf{y}_i, \hat{\mathbf{y}}_i^{(s)}\right) + \frac{\lambda_{\mathcal{T}}}{|\mathcal{T}|} \sum_{j \in \mathcal{T}} \ell\!\left(\mathbf{y}_j, \hat{\mathbf{y}}_j^{(s)}\right) + \lambda_{\mathbf{G}} \mathcal{L}^{(b)}_{\mathbf{G}}(\theta_1),
\end{equation}
where $\smash{\hat{\mathbf{y}}_i^{(s)} = f_{\theta_3}(\mathbf{z}_i^{(s)}), \, \hat{\mathbf{y}}_j^{(s)} = f_{\theta_3}(\mathbf{z}_j^{(s)})}$ and $\smash{\mathbf{z}_i^{(s)} \sim p(\mathbf{z}_i \, \lvert \, \mathbf{U}_\mathcal{C}, \mathbf{V}_\mathcal{C}), \, \mathbf{z}_j^{(s)} \sim p(\mathbf{z}_j \, \lvert \, \mathbf{U}_\mathcal{T}, \mathbf{U}_\mathcal{C}, \mathbf{V}_\mathcal{C})}$ are context-aware samples from Equations \ref{pzc} and \ref{pzt}, and $\lambda_{\mathcal{C}}$, $\lambda_{\mathcal{T}}$, $\lambda_{\mathbf{G}}$ are hyperparameters.

While the graph regularization term appears similar to knowledge distillation \citep{hinton2015distilling}, we emphasize that the former aims to preserve the covariance structures between the outputs of the image encoder $f_{\theta_1}$ rather than the outputs themselves. We believe that in light of new data, the image encoder should be able to update its potentially superficial representations of previously seen samples as long as it keeps the correlations between them unchanged. Indeed, some of the early regularization approaches based on knowledge distillation \citep{li2017learning,rebuffi2017icarl} are sometimes too restrictive and reportedly underperform in certain scenarios \citep{kemker2017fearnet}.

\smallskip
\textbf{Task-Free Knowledge Consolidation.} \enskip When task identities are not available, we use reservoir sampling \citep{vitter1985random} to update the context images and context labels as in \cite{riemer2018learning}. The sampling strategy takes as input a stream of data and randomly replaces a context sample in the episodic memory with a target sample, with probability proportional to the number of samples observed so far. Despite its simplicity, reservoir sampling has been shown to yield strong performance in recent work \citep{chaudhry2019continual,riemer2018learning,rolnick2019experience}.

While most prior work uses task boundaries to perform knowledge consolidation at the end of each task \citep{kirkpatrick2017overcoming,rebuffi2017icarl}, we update the context graph in memory after every minibatch of training data. In addition, such updates are performed at the sample level to maximize flexibility; we keep track of the cross entropy loss on each context sample and only update its edges in the graph when the model reaches a new low (denoted by $\smash{\mathcal{I}^{(b)}}$ previously). Intuitively, the loss measures how well the model has learned the context image through the connections it forms with others, so meaningful relations are most likely obtained at the bottom of the loss surface. Though samples from the same task often provide more support for each other, the task-agnostic mechanism for updating the context graph also allows for knowledge transfer across tasks when necessary.

\smallskip
\textbf{Memory and Time Complexity.} The inclusion of pairwise similarities and graph regularization result in a time and memory complexity of $\mathcal{O}(|\mathcal{M}|^2 + |\mathcal{M}|N)$ and $\mathcal{O}(|\mathcal{M}|^2)$, respectively, where $|\mathcal{M}|$ denotes the size of the episodic memory and $N$ the batch size for target images. The quadratic costs in $|\mathcal{M}|$, however, are not concerning in practice, as we deliberately use a small, fixed-size episodic memory. The cost of storing $\mathbf{G}$ is often dwarfed by the memory required for storing high-dimensional images, as each edge only needs one floating point number (see Appendix \ref{memory-usage} for more details on memory usage).

\section{Related Work} \label{sec:related-work}
\textbf{Continual Learning Approaches.} \enskip The existing work on continual learning mostly falls into three categories: \textit{regularization}, \textit{expansion}, and \textit{rehearsal}. \textit{Regularization} approaches alleviate catastrophic forgetting by penalizing changes in model weights that are important for past tasks. Different measures of weight importance are considered, including Fisher information \citep{kirkpatrick2017overcoming,chaudhry2018riemannian}, synaptic relevance \citep{zenke2017continual}, and uncertainty estimates \citep{ebrahimi2019uncertainty}. The constraints on weight updates can also be studied from Bayesian perspectives, where the posterior distribution of the weights is approximated and used as the prior for the next task \citep{nguyen2017variational,ritter2018online,titsias2019functional}. These regularization methods are efficient in memory and computational usage but suffer from brittleness due to representation drift \citep{titsias2019functional}.

\textit{Expansion} approaches dynamically allocate additional task-specific neural resources as more tasks arrive. \citet{rusu2016progressive}, for example, blocks changes to parameters learned for previous tasks and expands sub-networks while \citet{yoon2017lifelong} performs neuron splitting or duplication upon arrival of new tasks. Recently, non-parametric Bayesian approaches use Dirichlet process mixture models to expand a set of neural networks in a principled way \citep{jerfel2019reconciling,lee2020neural}. By design, these dynamic architectures prevent forgetting but quickly result in considerable model complexity.

Instead of growing model capacity, \textit{rehearsal} approaches maintain a small episodic memory of previous data or, alternatively, train a generative model to produce pseudo-data for past tasks, which are then replayed and interleaved with samples from the new task. Such generative models \citep{shin2017continual,kemker2017fearnet,achille2018life,caccia2019online,ostapenko2019learning} reduce working memory effectively, but they are also susceptible to catastrophic forgetting and invoke the complexity of the generative task \citep{parisi2019continual}. In contrast, episodic memory approaches are simpler and remarkably effective against forgetting \citep{rolnick2019experience,wu2019large}. \citet{lopez2017gradient} and \citet{chaudhry2018efficient}, for example, use an episodic storage of past data to impose inequality constraints on gradient updates while \citet{rebuffi2017icarl} constructs exemplars for knowledge distillation and nearest neighbor search. Recently, it has been shown that simple replay techniques and optimization-based meta-learning on the episodic memory outperform many previous approaches in online settings \citep{hayes2019memory,chaudhry2019continual,chaudhry2020using,riemer2018learning}. Our model is also based on experience replay, but it differs from the other approaches in the way the episodic memory is handled.

\smallskip
\textbf{Task-Free Continual Learning.} In real-world scenarios, task changes are often unknown and definitive boundaries between tasks do not always exist. However, most methods mentioned above rely on explicit task identities or task boundaries to consolidate knowledge or select sub-modules for task adaptation. Despite its significance, there are only a few works that address task-free continual learning. While \cite{aljundi2019task} heuristically detects peaks in the loss surface to consolidate knowledge, \cite{aljundi2019gradient,aljundi2019online} remove the need for task boundaries by a sample selection strategy for the episodic memory. Recently, the aforementioned non-parametric approaches train density estimators to detect task boundaries and perform model expansion \citep{lee2020neural,rao2019continual}. In contrast, our approach uses reservoir sampling \citep{vitter1985random} to update the episodic memory, similar to \cite{riemer2018learning,chaudhry2019continual}.

\smallskip
\textbf{Learning with Random Graphs.} \enskip Although widely studied in graph theory \citep{west2001introduction}, random graphs appear sparingly in the machine learning literature, perhaps more noticeably in neural architecture search \citep{xie2019exploring}. Our work is mostly related to previous work on functional neural process \citep{louizos2019functional}, where the authors build random graphs of dependencies to represent relational structures between context points in a stochastic process. Our approach is different in that (1) the random graphs are undirected and grow incrementally, (2) no variational inference is required, and (3) it addresses catastrophic forgetting and performs well under continual learning settings.

\begin{figure}[t]
    \centering
    \includegraphics[width=0.8\linewidth]{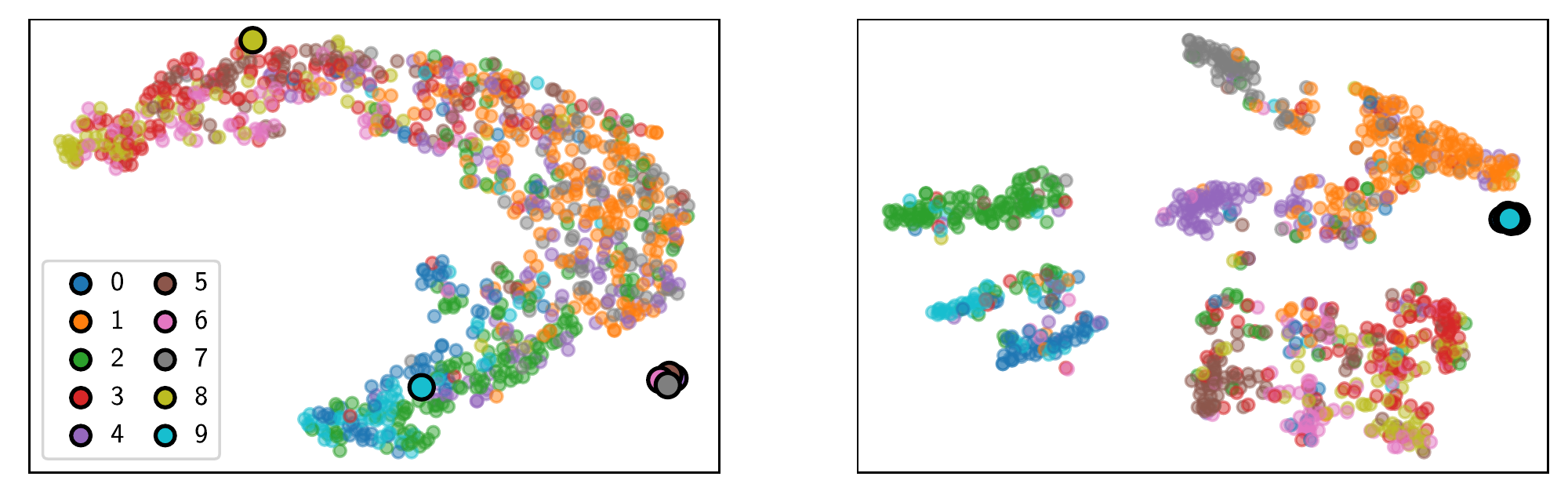}
    \caption{\small $t$-SNE visualization of image embeddings (small circles) from the penultimate layers and class embeddings (large circles) from the weights of the last layers on \textsc{Split SVHN}. The \textit{left} figure shows that \textbf{Finetune}, a model naively trained on the data stream, fails to recognize the class-based clustering structure and bias the image embeddings toward the last task (class 8 \& 9). In contrast, the \textit{right} figure shows that \textbf{GCL} (our model) maintains the relational structure and is more robust to the distributional shifts incurred by task changes.}
    \label{fig:embeddings}
\end{figure}

\smallskip
\textbf{Attention Mechanism.} \enskip While we motivate our approach from a graphical perspective, one can also consider it as some form of attention mechanism. In particular, the context graph $\mathbf{G}$ represents self-attention \citep{vaswani2017attention} across context images, and the context-target graph $\mathbf{A}$ represents cross-attention \citep{bahdanau2014neural} between context images and target images. Though advanced mechanisms such as multi-head attention have been applied successfully in many stationary settings \citep{vaswani2017attention,xu2015show,zhang2018self,kim2019attentive,sprechmann2018memory}, we note that naive applications of such techniques in online continual learning suffer from catastrophic forgetting due to representation drift when training switches to new tasks. In contrast, our model employs random attention, which arguably makes it more robust to such distributional shifts (see Figure \ref{fig:embeddings}).

\section{Experiments}
In this section, we evaluate the proposed GCL model on commonly used continual learning benchmarks. Additional results and details about the datasets, experiment setup, model architectures, and result analyses are available in the appendices.

\textbf{Experiment Setup.} \enskip We perform experiments on 6 image classification datasets: \textsc{Permuted MNIST}, \textsc{Rotated MNIST} \citep{lecun1998gradient}, \textsc{Split SVHN} \citep{netzer2011reading}, \textsc{Split CIFAR10} \citep{krizhevsky2009learning}, \textsc{Split CIFAR100} \citep{krizhevsky2009learning}, and \textsc{Split MiniImageNet} \citep{vinyals2016matching}. For each dataset, we follow \cite{lopez2017gradient,chaudhry2018efficient} and adopt the setting where the model only has access to an online stream of data with a batch size of 10 (see Appendix \ref{experiment-setup} for more details).

We consider both single-head and multiple-head settings. More specifically, we use single-head and one-epoch settings for our model and all baselines on \textsc{Permuted MNIST}, \textsc{Rotated MNIST}, \textsc{Split SVHN}, and \textsc{Split CIFAR10}. While most of previous work \citep{rebuffi2017icarl,lopez2017gradient,chaudhry2019continual} assume task identities on \textsc{Split CIFAR10}, we require all models to perform 10-way classification on each task with the same output head. This variant is more practical and challenging due to the need for incremental knowledge consolidation across tasks.

In addition, we also report results for multiple-head and 10-epochs settings on \textsc{Split CIFAR100} and \textsc{Split MiniImageNet}, following \cite{lopez2017gradient}. These datasets have more classes and fewer samples per class, rendering them too challenging for single-head settings.

\textbf{Model Architecture.} Our image encoders $f_{\theta_1}$ and $f_{\theta_2}$ partially share weights and are parametrized by an MLP on the MNIST variants and a simple 6-layer convolutional network on other datasets, each followed by a RELU activation and a separate linear mapping. As alluded earlier, we use an RBF kernel to compute similarities between image embeddings and find it sufficiently easy for initialization. The output mappings $f_{\theta_3}$ are MLPs in all cases (see Appendix \ref{model-architectures} for more details).

\smallskip
\textbf{Baselines.} \enskip We benchmark our model against multiple models, including (1) \textit{Finetune}, a popular baseline, naively trained on the data stream; (2) \textit{EWC} \citep{kirkpatrick2017overcoming}, an early regularization approach; (3) GEM \citep{lopez2017gradient}, a rehearsal approach based on an episodic memory of parameter gradients; (4) ER \citep{chaudhry2019continual}, a simple yet competitive experience method based on reservoir sampling; (5) MER \citep{riemer2018learning}, a rehearsal approach inspired by optimization-based meta-learning, and (6) ICARL \citep{rebuffi2017icarl} another well-known rehearsal strategy. Most of these baselines share the same model architectures: an MLP with two hidden layers on the MNIST variants, and a ResNet-18 \citep{he2016deep} on \textsc{Split SVHN} and \textsc{Split CIFAR10}, following \citep{lopez2017gradient} (see Appendix \ref{baseline-architectures} for more details).

\input{figures/permuted-cifar100}

\smallskip
\textbf{Metrics.} \enskip Following \citet{lopez2017gradient,chaudhry2018riemannian,chaudhry2019continual}, we evaluate the models using two classification metrics, namely, \textit{average accuracy} and \textit{average forgetting}:
\begin{equation}
	\label{eq:accuracy-forgetting}
	\textsc{ACC} \triangleq \frac{1}{T} \sum_{i = 1}^T R_{T, i}, \quad \textsc{FGT} \triangleq \frac{1}{T - 1} \sum_{j = 1}^{T - 1} (R_{T,i} - R_{i, i}),
\end{equation}
where $R_{i,j}$ denotes the test accuracy on task $j$ after the model has finished task $i$. Intuitively, the former measures the average test accuracy across all tasks while the latter measures the average decrease between each task's peak accuracy and its accuracy at the end of continual learning.

\begin{table}[t]
    \caption{\small Classification results ($\%$) on \textsc{Permuted MNIST}, \textsc{Rotated MNIST} and \textsc{Split SVHN}. The means and standard deviations are computed over five runs using different random seeds, When used, episodic memories contain 5 samples per class on average. The symbol $\uparrow$ ($\downarrow$) indicates that a higher (lower) number is better.}
    \label{permuted-rotated-mnist}
    \centering
    \begin{small}
        \begin{tabular}{l@{\hskip 0.15in}c@{\hskip 0.15in}c@{\hskip 0.15in}c@{\hskip 0.15in}c@{\hskip 0.15in}c@{\hskip 0.15in}c}
            \toprule
            \textsc{Dataset} & \multicolumn{2}{c}{\textsc{Permuted MNIST}} & \multicolumn{2}{c}{\textsc{Rotated MNIST}} & \multicolumn{2}{c}{\textsc{Split SVHN}}                                                                                        \\
            \cmidrule(r){1-1} \cmidrule(r){2-3} \cmidrule(r){4-5} \cmidrule(r){6-7}
            Method           & \textsc{ACC} $(\uparrow)$                   & \textsc{FGT} $(\downarrow)$                & \textsc{ACC} $(\uparrow)$               & \textsc{FGT} $(\downarrow)$ & \textsc{ACC} $(\uparrow)$ & \textsc{FGT}$(\downarrow)$ \\
            \midrule
            Finetune         & 60.19 $\pm$ 2.31                            & 23.62 $\pm$ 1.98                           & 43.80 $\pm$ 1.64                        & 46.52 $\pm$ 1.71            & 18.85 $\pm$ 0.10          & 94.78 $\pm$ 1.24           \\
            EWC              & 64.94 $\pm$ 1.22                            & 18.33 $\pm$ 1.07                           & 44.99 $\pm$ 1.73                        & 44.98 $\pm$ 1.95            & 18.76 $\pm$ 0.27          & 94.99 $\pm$ 1.23           \\
            \midrule
            GEM              & 79.17 $\pm$ 0.70                            & 3.68 $\pm$ 0.68                            & 82.60 $\pm$ 0.48                        & 5.47 $\pm$ 0.45             & 33.40 $\pm$ 3.27          & 68.91 $\pm$ 4.06           \\
            ER               & 79.90 $\pm$ 0.46                            & 3.78 $\pm$ 0.45                            & 80.82 $\pm$ 0.68                        & 6.78 $\pm$ 0.69             & 45.41 $\pm$ 3.03          & 62.37 $\pm$ 4.33           \\
            MER              & 79.68 $\pm$ 0.42                            & 3.47 $\pm$ 0.41                            & 83.56 $\pm$ 0.23                        & 8.14 $\pm$ 0.46             & -                         & -                          \\
            \textbf{GCL}     & \textbf{82.36} $\pm$ 0.36                   & \textbf{2.92} $\pm$ 0.23                   & \textbf{86.37} $\pm$ 0.32               & \textbf{3.22} $\pm$ 0.50    & \textbf{60.68} $\pm$ 1.67 & \textbf{21.86} $\pm$ 2.35  \\
            \bottomrule
        \end{tabular}
    \end{small}
    \label{table:mnist}
\end{table}

\pgfplotstableread{
    task ftm fts ewcm ewcs gemm gems erm ers gclm gcls
    1 93.020 1.065 93.020 1.065 90.820 2.504 93.530 0.544 95.550 0.322
    2 37.560 2.089 38.465 1.382 51.940 3.978 60.185 3.207 73.905 1.084
    3 28.440 0.889 28.483 0.649 34.113 5.231 40.723 4.220 55.400 1.719
    4 23.110 0.537 23.103 0.554 27.207 3.966 32.818 1.831 52.655 3.100
    5 18.460 0.116 18.492 0.131 22.880 3.413 29.942 3.080 49.616 1.850
}\splitcifar

\pgfplotstableread{
    task ftm fts ewcm ewcs icarlm icarls gemm gems erm ers gclm gcls
    1 66.120 4.313 68.840 0.833 41.720 3.563 51.920 3.557 44.259 10.073 75.856 3.199
    2 60.760 2.197 62.240 2.746 34.480 3.978 46.820 5.912 65.805 2.510 89.951 0.268
    3 58.053 1.896 57.840 0.981 35.667 3.790 47.933 3.152 61.850 2.426 91.900 0.378
    4 56.210 0.885 54.350 1.371 39.560 2.139 46.120 4.090 57.455 5.534 91.813 0.281
    5 57.088 1.701 56.496 1.029 42.240 3.898 50.720 3.152 55.215 8.169 91.651 0.218
    6 53.427 1.653 53.900 1.185 40.307 3.465 50.007 1.843 55.201 7.379 91.701 0.113
    7 53.663 1.611 53.006 2.489 42.697 2.674 51.314 1.162 60.272 0.825 91.273 0.128
    8 55.565 1.687 54.200 0.848 44.835 1.725 51.870 0.764 55.837 5.752 91.445 0.083
    9 52.302 2.122 53.956 1.178 43.778 0.719 50.911 1.320 58.785 2.059 90.784 0.128
    10 54.808 1.819 54.960 1.226 46.472 2.015 52.300 1.172 54.052 1.405 90.277 0.219
    11 53.938 1.691 53.524 1.454 46.055 3.756 53.425 0.938 55.733 2.008 89.943 0.290
    12 52.810 1.162 52.873 2.735 46.837 2.708 54.253 1.526 54.011 2.010 89.454 0.329
    13 54.948 1.530 55.828 1.307 48.683 2.946 55.123 1.042 47.186 11.763 89.449 0.281
    14 53.409 2.403 53.406 0.756 48.540 2.244 54.594 0.959 52.121 2.012 89.084 0.347
    15 53.792 2.350 53.664 2.203 51.245 1.990 55.592 1.128 55.561 1.088 88.647 0.275
    16 52.752 2.216 53.292 1.514 51.395 1.284 55.297 0.836 57.729 2.320 88.166 0.376
    17 54.729 1.991 54.122 1.713 51.111 1.689 56.419 0.701 56.672 2.306 87.645 0.335
    18 53.362 1.939 53.484 1.456 50.887 1.058 56.209 0.986 55.973 1.840 87.526 0.253
    19 55.587 2.134 55.815 1.228 50.956 0.846 57.255 1.127 58.571 0.917 87.225 0.358
    20 55.390 1.943 55.604 1.114 52.180 0.269 57.898 1.127 58.743 0.738 86.370 0.325
}\splitimagenet

\begin{table}[t]
    \caption{\small Classification results ($\%$) on \textsc{Split CIFAR10} and \textsc{Split CIFAR100} and \textsc{Split MiniImageNet}. The means and standard deviations are computed over five runs using different random seeds, When used, episodic memories contain 5 samples per class on average. The symbol $\uparrow$ ($\downarrow$) indicates that a higher (lower) number is better.}
    \centering
    \begin{small}
        \begin{tabular}{l@{\hskip 0.15in}c@{\hskip 0.15in}c@{\hskip 0.15in}c@{\hskip 0.15in}c@{\hskip 0.15in}c@{\hskip 0.15in}c}
            \toprule
            \textsc{Dataset} & \multicolumn{2}{c}{\textsc{Split CIFAR10}} & \multicolumn{2}{c}{\textsc{Split CIFAR100}} & \multicolumn{2}{c}{\textsc{Split MiniImageNet}}                                                                                         \\
            \cmidrule(r){1-1} \cmidrule(r){2-3} \cmidrule(r){4-5} \cmidrule(r){6-7}
            Method           & \textsc{ACC} $(\uparrow)$                  & \textsc{FGT} $(\downarrow)$                 & \textsc{ACC} $(\uparrow)$                       & \textsc{FGT} $(\downarrow)$ & \textsc{ACC} $(\uparrow)$ & \textsc{FGT} $(\downarrow)$ \\
            \midrule
            Finetune         & 18.46 $\pm$ 0.12                           & 86.48 $\pm$ 1.02                            & 55.39 $\pm$ 1.94                                & 25.94 $\pm$ 1.89            & 37.84 $\pm$ 0.87          & 31.41 $\pm$ 1.57            \\
            EWC              & 18.49 $\pm$ 0.13                           & 86.95 $\pm$ 1.15                            & 55.60 $\pm$ 1.11                                & 23.53 $\pm$ 1.19            & 36.61 $\pm$ 2.06          & 28.17 $\pm$ 4.49            \\
            \midrule
            ICARL            & -                                          & -                                           & 58.08 $\pm$ 1.44                                & 24.22 $\pm$ 1.35            & -                         & -                           \\
            GEM              & 22.88 $\pm$ 3.41                           & 76.90 $\pm$ 5.53                            & 65.66 $\pm$ 0.70                                & 15.52 $\pm$ 0.41            & 54.06 $\pm$ 0.22          & 13.17 $\pm$ 0.74            \\
            ER               & 29.94 $\pm$ 3.08                           & 72.64 $\pm$ 4.88                            & 69.40 $\pm$ 1.21                                & 11.25 $\pm$ 1.24            & 58.74 $\pm$ 0.74          & 9.02 $\pm$ 2.49             \\
            \textbf{GCL}     & \textbf{49.62} $\pm$ 1.85                  & \textbf{35.69} $\pm$ 3.33                   & \textbf{74.51} $\pm$ 0.99                       & \textbf{6.54} $\pm$ 1.26    & \textbf{61.54} $\pm$ 0.57 & \textbf{6.10} $\pm$ 2.73    \\
            \bottomrule
        \end{tabular}
    \end{small}
    \label{table:svhn-cifar10-1}
\end{table}

\pgfplotstableread{
    memory svhnerm svhners svhngclm svhngcls cifarerm cifarers cifargclm cifargcls
    100 31.12 2.21 45.79 5.92 21.54 0.83 40.01 2.33
    250 45.51 3.03 60.68 1.67 29.94 3.08 49.62 1.85
    500 57.51 2.77 65.79 1.54 36.08 1.09 53.87 0.97
    1000 67.33 1.66 69.95 0.78 45.75 1.82 57.26 0.28
}\memoryaccuracy

\begin{minipage}[b]{0.45\linewidth}
    \centering
    \begin{figure}[H]
        \centering
        \begin{tikzpicture}[declare function={barWidth=8pt; barShift=barWidth/2;}]
            \pgfplotsset{
                xtick=data, major x tick style=transparent, ticklabel style={font=\scriptsize\sffamily\sansmath, yshift=3pt},
                every axis label/.append style={font=\sffamily\sansmath\scriptsize},
                legend style={font=\scriptsize\sffamily, legend columns=-1, /tikz/every even column/.append style={column sep=0.5cm}},
            }
            \begin{axis}[
                    height=5cm, width=\linewidth,
                    ybar=0pt, bar width=barWidth,
                    xmin=100, xmax=1000, symbolic x coords={100, 250, 500, 1000}, enlarge x limits=0.1, xtick=data,
                    ymin=0, ymax=75,
                    ymajorgrids=true, ytick distance=25,
                    legend image code/.code={\draw[#1, draw=black] (0cm,-0.1cm) rectangle (0.6cm,0.1cm);},
                    title=\textsc{Split CIFAR10},
                    xlabel=Memory Size, xlabel style={yshift=2pt},
                    ylabel=Average Accuracy (\%),
                    enlarge x limits=0.15,
                ]
                \addplot+[mark=*, mark options={xshift=-barShift, mark size=1.5pt}, draw=black, fill=knred!80,
                    error bars/.cd, y dir=both, y explicit, error bar style={line width=1pt,solid, black},] table[x=memory, y=cifarerm, y error=cifarers] {\memoryaccuracy};
                \addplot+[mark=*, mark options={xshift=barShift, mark size=1.5pt}, draw=black, fill=kngreen!80,
                    error bars/.cd, y dir=both, y explicit, error bar style={line width=1pt,solid, black},] table[x=memory, y=cifargclm, y error=cifargcls] {\memoryaccuracy};
                \legend{ER, \textbf{GCL}}
            \end{axis}
        \end{tikzpicture}
        \caption{\small Effects of episodic memory sizes.}
        \label{fig:memory-cifar10}
    \end{figure}
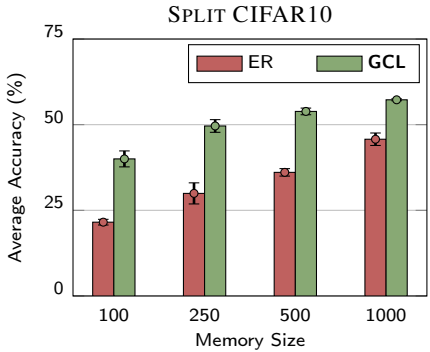
\end{minipage}\quad
\begin{minipage}[b]{0.45\linewidth}
    \centering
    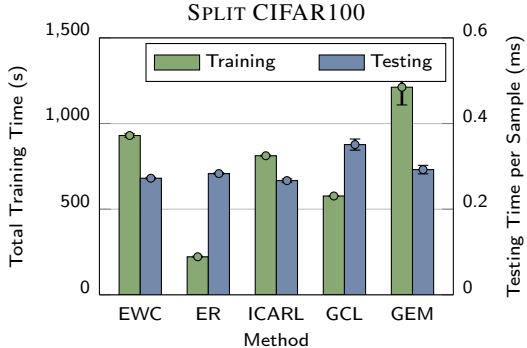
\begin{figure}[H]
        \centering
        \begin{tikzpicture}[declare function={barWidth=8pt; barShift=barWidth/2;}]
            \pgfplotsset{
                xtick=data, major x tick style=transparent, title style={font=\footnotesize},
                ticklabel style={font=\scriptsize\sffamily\sansmath, yshift=3pt},
                legend style={font=\scriptsize\sffamily, legend columns=-1, /tikz/every even column/.append style={column sep=0.5cm}},
            }
            \begin{axis}[
                    height=5cm, width=\linewidth,
                    ybar, bar width=barWidth, bar shift=-barShift,
                    symbolic x coords={EWC, ER, ICARL, GCL, GEM}, enlarge x limits=0.15,
                    axis y line*=left, axis x line=none,
                    ymin=0, ymax=1500,
                    ymajorgrids=true, ytick distance=500,
                    ylabel=Total Training Time (s),
                    title=\textsc{Split CIFAR100},
                ]
                \addplot[
                    mark=*, mark options={xshift=-barShift, mark size=1.5pt}, draw=black, fill=kngreen!80,
                    error bars/.cd, y dir=both, y explicit, error bar style={line width=1pt,solid, black},
                ] coordinates {
                        (EWC, 930) +- (0, 10.149)
                        (ICARL, 812) +- (0, 3)
                        (GEM, 1212) +- (0, 103.56)
                        (ER, 222) +- (0, 4)
                        (GCL, 577) +- (0, 2.64)
                    };
                \label{train-time}
            \end{axis}

            \begin{axis}[
                    height=5cm, width=\linewidth,
                    ybar, bar width=barWidth, bar shift=barShift,
                    symbolic x coords={EWC, ER, ICARL, GCL, GEM}, enlarge x limits=0.15,
                    axis y line*=right,
                    ymin=0, ymax=0.6, ytick distance=0.2,
                    legend image code/.code={\draw[#1, draw=black] (0cm,-0.1cm) rectangle (0.6cm,0.1cm);},
                    ylabel=Testing Time per Sample (ms),
                    xlabel=Method, xlabel style={yshift=2pt},
                ]
                \addlegendimage{/pgfplots/refstyle=graph-barplot}
                \addlegendentry{Training}
                \addplot[
                    mark=*, mark options={xshift=barShift, mark size=1.5pt}, fill=knblue!70,
                    error bars/.cd, y dir=both, y explicit, error bar style={line width=1pt,solid, black}
                ] coordinates {
                        (EWC, 0.2723) +- (0, 0.002)
                        (ICARL, 0.2667) +- (0, 0.004)
                        (GEM, 0.2924) +- (0, 0.01)
                        (ER, 0.2831) +- (0, 0.002)
                        (GCL, 0.3508) +- (0, 0.013)
                    };
                \addlegendentry{Testing}
            \end{axis}
        \end{tikzpicture}
        \caption{\small Training and testing time.}
        \label{fig:train-test}
    \end{figure}
\end{minipage}
\vspace{0.5cm}

\textbf{Classification Performance.} \enskip
Table \ref{table:mnist} and \ref{table:svhn-cifar10-1} show the overall experimental results, and the evolution of performance as a function of the number of tasks are detailed in Figure \ref{fig:accuracy-tasks-1}. In every setting, our model (GCL) outperforms the baselines by significant margins, and the gains in performance are especially substantial on complex datasets such as \textsc{Split CIFAR10} or \textsc{Split CIFAR100}. As noted by \cite{chaudhry2018efficient}, EWC \citep{kirkpatrick2017overcoming} performs poorly without multiple passes over the datasets, and we additional find that GEM \citep{lopez2017gradient} is not very effective under the single-head variants (e.g. on \textsc{Split SVHN} or \textsc{Split CIFAR10}). Task-free approaches such as ER and MER perform more favorably, and such findings are consistent with recent studies \citep{chaudhry2019continual,riemer2018learning}.

The advantageous performance of GCL over the other rehearsal strategies can be attributed to its efficient use of the episodic memory. Figure \ref{fig:memory-cifar10} shows that both ER \citep{chaudhry2019continual} and GCL benefit from increases in memory size, but the outperformance of GCL is more visible under the low-resource regime. Sample efficiency, as demonstrated, is especially important since the memory constraints are not relaxable despite the growing complexity of the data distribution during training. It is also worth emphasizing that although our model takes more time to train and evaluate at test time than ER, its training time and testing time are comparable to other approaches (see Figure \ref{fig:train-test}).

\textbf{Learned Graphs.} \enskip Central to our approach are the pairwise similarities between context images captured by the context graph $\mathbf{G}$. Figure \ref{fig:context-graph} shows a continuous realization of the context graph at the end of continual learning on \textsc{Split CIFAR10}, which has been sorted according to context labels placed underneath the adjacency matrix. Despite being trained exclusively on two classes of target images at a time (e.g., plane \& car or bird \& cat), the model appears to learn the clustering structure of images relatively well with more pronounced edges formed within classes than across them. The edges across tasks are noisier, but some edges indicate intuitive visual similarities such as those between images of car and truck. We note that the 10-way classification setup in each task encourages the model to clear inter-class edges, especially those within each binary task, so the degree of knowledge transfer across tasks is understandably more subtle.

\input{figures/context-regularization}

\begin{wraptable}{r}{8cm}
    \vspace{-0.1cm}
    \caption{\small Ablation study on \textsc{Split CIFAR10}.}
    \label{ablation-results}
    \centering
    \begin{small}
        \begin{tabular}{lccccc}
            \toprule
            Graph regularization                       & \checkmark     & $\times$   & $\times$   & $\times$   \\
            Multiple graph samples                     & \checkmark     & \checkmark & $\times$   & $\times$   \\
            Random $\mathbf{G}$ \& $\mathbf{A}$        & \checkmark     & \checkmark & \checkmark & $\times$   \\
            Deterministic $\mathbf{G}$ \& $\mathbf{A}$ & $\times$       & $\times$   & $\times$   & \checkmark \\
            \midrule
            Average accuracy                           & \textbf{49.62} & 44.04      & 42.08      & 30.50      \\
            \bottomrule
        \end{tabular}
    \end{small}
    \label{table:ablation}
\end{wraptable}

\smallskip
\textbf{Ablation Study.} We further investigate our model performance with an ablation study and summarize it in Table \ref{table:ablation}. Without the graph regularization term in Equation \ref{eq:graph-regularization}, the model significantly performs worse, indicating that past connections between context samples can help alleviate catastrophic forgetting. By varying the hyper-parameter $\lambda_{\mathbf{G}}$, we also see from Figure \ref{fig:graph-regularization} that an extreme amount of graph regularization (e.g. $\lambda_{\mathbf{G}} = 1000$) can have detrimental effects on the model performance as well. As alluded earlier, the ability to draw multiple graph samples and average their predictions at test time brings out some gains, as often the case with ensemble methods. Perhaps more importantly, we find that making the context graph $\mathbf{G}$ and the context-target graph $\mathbf{A}$ deterministic results in a dramatic drop in accuracy. The resulting model is a variant of attention mechanism, most similar to attentive neural process \citep{kim2019attentive}, and as discussed in Section \ref{sec:related-work}, such a deterministic model often relies on a handful of edges, all of which are also prone to distributional shifts and thus catastrophic forgetting as well.

\section{Conclusion and Discussion}
In this work, we have introduced a graph-based approach to continual learning that exploits pairwise similarities between samples to support knowledge transfer. Based on the learned graphs, we derive a regularization term to guide the training of new tasks against catastrophic forgetting. Our model demonstrates an efficient use of the episodic memory, and as a result, performs competitively under various settings, without requiring access to task definition both during training and at test time in some cases.

As graph-based approaches, including ours, offer a natural way to describe relational inductive biases \citep{battaglia2018relational}, we hope that future works further examine the applications of graphs under continual learning settings. If trained well, these graphs can be used not only to share knowledge but also to minimize inference between samples and tasks. A promising direction, for example, is to pose the problem of updating the episodic memory as a graph search and leverage the rich literature on graph theory to devise better  strategies for sample selection. As demonstrated by previous works \citep{aljundi2019gradient,isele2018selective}, such selection mechanisms can be effective against catastrophic forgetting, especially when the data distribution is not balanced across tasks.

\section{Acknowledgment}
Financial support is gratefully acknowledged from a Xerox PARC Faculty Research Award, National Science Foundation Awards 1455172, 1934985, 1940124, and 1940276, USAID, and Cornell University Atkinson Center for a Sustainable Future.

\bibliography{references}
\bibliographystyle{iclr2021_conference}

\newpage

\appendix
\section{Experiment Setup} \label{experiment-setup}
We perform experiments on six commonly used classification datasets: \textsc{Permuted MNIST}, \textsc{Rotated MNIST} \citep{lecun1998gradient}, \textsc{Split SVHN} \citep{netzer2011reading}, \textsc{Split CIFAR10} \citep{krizhevsky2009learning}, \textsc{Split CIFAR100} \citep{krizhevsky2009learning}, and \textsc{Split MiniImageNet} \citep{vinyals2016matching}.
\begin{itemize}
	\item \textsc{Permuted MNIST} \citep{goodfellow2013empirical} is a variant of the \textsc{MNIST} dataset of handwritten digits \citep{lecun1998gradient}, where each task applies a fixed random pixel permutation to the original dataset. The benchmark dataset consists of 20 tasks, each with 1000 samples from 10 different classes.

	\item \textsc{Rotated MNIST} \citep{lopez2017gradient} is another variant of the \textsc{MNIST} dataset of handwritten digits \citep{lecun1998gradient}, where each task applies a fixed random image rotation to the original dataset. The benchmark dataset consists of 20 tasks, each with 1000 samples from 10 different classes.

	\item \textsc{Split SVHN} is a variant of the \textsc{SVHN} dataset \citep{netzer2011reading} that consists of 5 tasks, each with two consecutive classes. Since the benchmark dataset is much more challenging than the MNIST variants, we use all of its 73,257 training samples (i.e. 14,650 samples per task) to train our model and the baselines.

	\item \textsc{Split CIFAR10} is a variant of the \textsc{CIFAR-10} dataset \citep{krizhevsky2009learning}. Similar \textsc{Split SVHN}, the benchmark dataset consists of 5 tasks, each with two consecutive classes. We use all of its 50,000 training samples (i.e. 10,000 samples per task) to train our model and the baselines.

	\item \textsc{Split CIFAR100} is a variant of the \textsc{CIFAR-100} dataset \citep{krizhevsky2009learning}. The benchmark dataset consists of 20 tasks, each with 5 consecutive classes. We use all of its 50,000 training samples (i.e. 2,500 samples per task) to train our model and the baselines.

	\item \textsc{Split MiniImageNet} is a variant of the \textsc{MiniImageNet} dataset \citep{krizhevsky2009learning}. The benchmark dataset consists of 20 tasks, each with 5 consecutive classes. We use all of its 50,000 training samples (i.e. 2,500 samples per task) to train our model and the baselines. Each image is resized to 84 $\times$ 84 pixels.
\end{itemize}

\section{Model Architectures} \label{model-architectures}
As mentioned, while most of previous work uses multi-head architectures and assumes knowledge of task boundaries at test time, we employ a shared classifier head for all tasks. For the MNIST datasets, the image encoders $f_{\theta_1}$ (for graph construction) and $f_{\theta_2}$ (for latent computation) share a multi-layered perceptron with two hidden layers of 256 ReLU neurons, followed by two separate linear mappings, one for each of the encoders. For \textsc{Split SVHN}, \textsc{Split CIFAR10}, \textsc{Split CIFAR100}, and \textsc{Split MiniImageNet}, the image encoders share a simple convolutional network with the following structure: \texttt{conv 64} $\rightarrow$ \texttt{conv 64} $\rightarrow$ \texttt{maxpool} $\rightarrow$ \texttt{conv 64} $\rightarrow$ \texttt{conv 64} $\rightarrow$ \texttt{maxpool} $\rightarrow$ \texttt{conv 64} $\rightarrow$ \texttt{conv 64} $\rightarrow$ \texttt{maxpool}, where \texttt{conv NF} is a $3 \times 3$ convolution with \texttt{NF} output filters, BatchNorm, and ReLU activations. For all datasets, another linear mapping follows the image encoder $f_{\theta_1}$ before a Gaussian kernel computes the similarities between image embeddings. Finally, the classifier head consists of a RELU activation and a single linear mapping.

\section{Baseline Architectures} \label{baseline-architectures}
We use the same neural network architectures for all the baselines described in the paper: a multi-layered perceptron with two hidden layers of 400 ReLU neurons on \textsc{Permuted MNIST} and \textsc{Rotated MNIST}, following \citep{hsu2018re}, and a ResNet-18 \citep{he2016deep} with 20 filters across all layers on other datasets, following \citep{lopez2017gradient}. For all datasets, the baselines consist of more parameters than our corresponding models (see Table \ref{model-sizes} for more details).

We adopt the implementations of EWC \citep{kirkpatrick2017overcoming}, GEM \citep{lopez2017gradient}, and MER \citep{riemer2018learning} from the authors' repositories \footnote{https://github.com/facebookresearch/GradientEpisodicMemory} \footnote{https://github.com/mattriemer/mer}.

\begin{table}[t]
    \caption{\small Number of trainable parameters in continual learning models.}
    \label{model-sizes}
    \centering
    \begin{small}
        \begin{tabular}{lcccccc}
            \toprule
            Method                      & Finetune & EWC   & GEM   & ER    & MER  & \textbf{GCL} \\
            \midrule
            \textsc{Split MNIST}        & 478K     & 478K  & 478K  & 478K  & 478K & 406K         \\
            \textsc{Permuted MNIST}     & 478K     & 478K  & 478K  & 478K  & 478K & 406K         \\
            \textsc{Rotated MNIST}      & 478K     & 478K  & 478K  & 478K  & 478K & 406K         \\
            \midrule
            \textsc{Split SVHN}         & 1.09M    & 1.09M & 1.09M & 1.09M & -    & 326K         \\
            \textsc{Split CIFAR10}      & 1.09M    & 1.09M & 1.09M & 1.09M & -    & 326K         \\
            \textsc{Split CIFAR100}     & 1.09M    & 1.09M & 1.09M & 1.09M & -    & 326K         \\
            \textsc{Split MiniImageNet} & 1.09M    & 1.09M & 1.09M & 1.09M & -    & 343K         \\
            \bottomrule
        \end{tabular}
    \end{small}
\end{table}

\section{Additional Task-Free Baselines}
We also note that despite our attempts to tune parameters for MER \citep{riemer2018learning} on  \textsc{Split SVHN} and \textsc{Split CIFAR10}, the baseline does not perform reasonably well. The model uses a batch size of 1 and requires multiple passes through the episodic memory per batch, so it is much slower than our model and all other baselines. Due to limited time and computational resources, we do not further investigate the baseline and therefore avoid reporting immature results for fairness.

However, we include results of \textsc{CN-DPM} \citep{lee2020neural}, a competitive task-free model based on Dirichlet process mixture models in Table \ref{cndpm-results}. Our setup for \textsc{Split CIFAR10} is analogous to that of \cite{lee2020neural}, so we directly quote the numbers for \textsc{CN-DPM} from the paper. Although \textsc{CN-DPM} performs favorably among task-free approaches to continually learning, including GSS \citep{aljundi2019gradient}, our model outperforms \textsc{CN-DPM} by a significant margin, even when using a smaller memory size.

\begin{table}[H]
    \caption{\small \textsc{GCL} results and \textsc{CN-DPM} results with different memory sizes.}
    \label{cndpm-results}
    \centering
    \begin{small}
        \begin{tabular}{lcccc}
            \toprule
            \multirow{2}{*}{Method}          & \multicolumn{2}{c}{\textsc{Split SVHN}} & \multicolumn{2}{c}{\textsc{Split CIFAR10}}                                                         \\
            \cmidrule(r){2-3} \cmidrule(r){4-5}
                                             & 250                                     & 500                                        & 500                       & 1000                      \\
            \midrule
            ER \citep{chaudhry2019continual} & 45.51 $\pm$ 3.03                        & 57.51 $\pm$ 2.77                           & 36.08 $\pm$ 1.09          & 45.75 $\pm$ 1.82          \\
            CN-DPM \citep{lee2020neural}     & $-$                                     & $-$                                        & 43.07 $\pm$ 0.16          & 45.21 $\pm$ 0.18          \\
            \textbf{GCL} (Ours)              & \textbf{60.68} $\pm$ 1.67               & \textbf{65.79} $\pm$ 1.54                  & \textbf{53.87} $\pm$ 0.97 & \textbf{57.26} $\pm$ 0.28 \\
            \bottomrule
        \end{tabular}
    \end{small}
\end{table}

\section{Memory Usage} \label{memory-usage}
Both GCL and ER \citep{chaudhry2019continual} uses an episodic memory to store images and labels from past tasks. The only additional memory usage of GCL comes from the context graph $\mathbf{G}$, which is represented by a square matrix whose entries intuitively describe pairwise similarities between such images. Given a memory consisting of $|\mathcal{M}|$ images of size $C \times H \times W$, it only requires $|\mathcal{M}|^2$ floating points to store the matrix.

\begin{table}[H]
    \caption{\small Memory usage of ER and GCL for various datasets.}
    \centering
    \begin{small}
        \begin{tabular}{lrrrr}
            \toprule
            \textsc{Dataset}            & $|\mathcal{M}|$ & Image Size                & ER        & GCL       \\
            \midrule
            \textsc{Permuted MNIST}     & 1000            & 1 $\times$ 28 $\times$ 28 & 3.284 MB  & 7.284 MB  \\
            \textsc{Rotated MNIST}      & 1000            & 1 $\times$ 28 $\times$ 28 & 3.284 MB  & 7.284 MB  \\
            \textsc{Split CIFAR10}      & 250             & 3 $\times$ 32 $\times$ 32 & 3.109 MB  & 3.359 MB  \\
            \textsc{Split SVHN}         & 250             & 3 $\times$ 32 $\times$ 32 & 3.109 MB  & 3.359 MB  \\
            \textsc{Split CIFAR100}     & 500             & 3 $\times$ 32 $\times$ 32 & 6.219 MB  & 7.199 MB  \\
            \textsc{Split MiniImageNet} & 500             & 3 $\times$ 84 $\times$ 84 & 42.408 MB & 43.389 MB \\
            \bottomrule
        \end{tabular}
    \end{small}
    \label{table:memory-calulation}
\end{table}

As seen from Table \ref{table:memory-calulation}, the memory usage of GCL are very similar the same as that of ER, except when both are very small as in the case of \textsc{Permuted MNIST} and \textsc{Rotated MNIST}, because (1) continual learning algorithms are often required to use a very small $|\mathcal{M}|$ and (2) the cost for storing natural images are often much higher than that of the context graph.

As the number of tasks increases, it is perhaps essential to expand the episodic memory, in which case the quadratic growth of the latter might dominate the linear increase of the former (e.g. $|\mathcal{M}| = 5000$ and images are of size $3 \times 32 \times 32$). Although we have not practically encountered such a problem with GCL, we note that the quadratic growth of the number of entries in the context graph can be reduced to a linear growth in memory requirements. More specifically, each entry is the output of the kernel function $\kappa_\tau$ (see Section 3, e.g. $\kappa_\tau(\mathbf{u}_i, \mathbf{u}_j) = \smash{\exp\left(-\frac{\tau}{2}\|\mathbf{u}_i - \mathbf{u}_j\|_2^2  \right)}$), so we could easily store $|\mathcal{M}|$ intermediate embeddings $\{\mathbf{u}_i\}$ at each step and apply the kernel function on the fly, which is especially beneficial when $\mathbf{u}_i$ are much lower dimensional than the original images.

\section{Additional Experiment Results}

\input{figures/accuracy-task}

\begin{figure}[H]
    \centering
    \begin{minipage}{.49\textwidth}
        \begin{tikzpicture}[declare function={barWidth=10pt; barShift=barWidth/2;}]
            \pgfplotsset{
                xtick=data, major x tick style=transparent, ticklabel style={font=\scriptsize\sffamily\sansmath, yshift=3pt},
                every axis label/.append style={font=\sffamily\sansmath\scriptsize},
                legend style={font=\scriptsize\sffamily, legend columns=-1, /tikz/every even column/.append style={column sep=0.5cm}},
            }
            \begin{axis}[
                    height=5cm, width=\linewidth,
                    ybar=0pt, bar width=barWidth,
                    xmin=100, xmax=1000, symbolic x coords={100, 250, 500, 1000}, enlarge x limits=0.1, xtick=data,
                    ymin=0, ymax=75,
                    ymajorgrids=true, ytick distance=25,
                    legend image code/.code={\draw[#1, draw=black] (0cm,-0.1cm) rectangle (0.6cm,0.1cm);},
                    title=\textsc{Split SVHN},
                    ylabel=Average Accuracy (\%),
                    xlabel=Memory Size, xlabel style={yshift=2pt},
                    enlarge x limits=0.15,
                ]
                \addplot+[mark=*, mark options={xshift=-barShift, mark size=1.5pt}, draw=black, fill=knred!80,
                    error bars/.cd, y dir=both, y explicit, error bar style={line width=1pt,solid, black},] table[x=memory, y=svhnerm, y error=svhners] {\memoryaccuracy};
                \addplot+[mark=*, mark options={xshift=barShift, mark size=1.5pt}, draw=black, fill=kngreen!80,
                    error bars/.cd, y dir=both, y explicit, error bar style={line width=1pt,solid, black},] table[x=memory, y=svhngclm, y error=svhngcls] {\memoryaccuracy};
            \end{axis}
        \end{tikzpicture}
    \end{minipage}\hspace{0.1cm}
    \begin{minipage}{.49\textwidth}
        \begin{tikzpicture}[declare function={barWidth=10pt; barShift=barWidth/2;}]
            \pgfplotsset{
                xtick=data, major x tick style=transparent, ticklabel style={font=\scriptsize\sffamily\sansmath, yshift=3pt},
                every axis label/.append style={font=\sffamily\sansmath\scriptsize},
                legend style={font=\scriptsize\sffamily, legend columns=-1, /tikz/every even column/.append style={column sep=0.5cm}},
            }
            \begin{axis}[
                    height=5cm, width=\linewidth,
                    ybar=0pt, bar width=barWidth,
                    xmin=100, xmax=1000, symbolic x coords={100, 250, 500, 1000}, enlarge x limits=0.1, xtick=data,
                    ymin=0, ymax=75,
                    ymajorgrids=true, ytick distance=25,
                    legend image code/.code={\draw[#1, draw=black] (0cm,-0.1cm) rectangle (0.6cm,0.1cm);},
                    title=\textsc{Split CIFAR10},
                    xlabel=Memory Size, xlabel style={yshift=2pt},
                    enlarge x limits=0.15,
                ]
                \addplot+[mark=*, mark options={xshift=-barShift, mark size=1.5pt}, draw=black, fill=knred!80,
                    error bars/.cd, y dir=both, y explicit, error bar style={line width=1pt,solid, black},] table[x=memory, y=cifarerm, y error=cifarers] {\memoryaccuracy};
                \addplot+[mark=*, mark options={xshift=barShift, mark size=1.5pt}, draw=black, fill=kngreen!80,
                    error bars/.cd, y dir=both, y explicit, error bar style={line width=1pt,solid, black},] table[x=memory, y=cifargclm, y error=cifargcls] {\memoryaccuracy};
                \legend{ER, \textbf{GCL}}
            \end{axis}
        \end{tikzpicture}
    \end{minipage}
    \caption{\small Average accuracy as a function of memory size on \textsc{Split SVHN} and \textsc{Split CIFAR10}.}
    \label{fig:memory-svhn-cifar}
\end{figure}
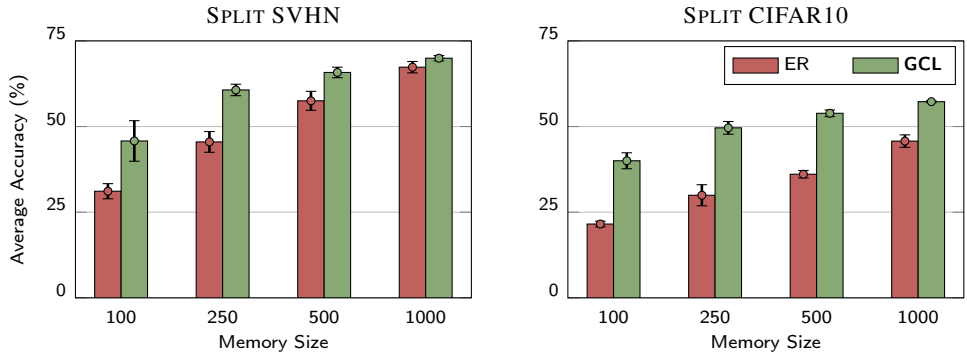

\pgfplotstableread{
    memory svhnerm svhners svhngclm svhngcls cifarerm cifarers cifargclm cifargcls
    100 79.76 2.80 25.49 5.84 84.59 1.38 42.30 3.79
    250 62.37 4.33 21.68 2.35 72.64 4.88 35.69 3.33
    500 46.79 4.32 16.65 1.08 64.58 1.30 31.18 1.55
    1000 33.88 1.39 15.37 1.22 52.15 2.56 28.53 0.68
}\memoryforgetting

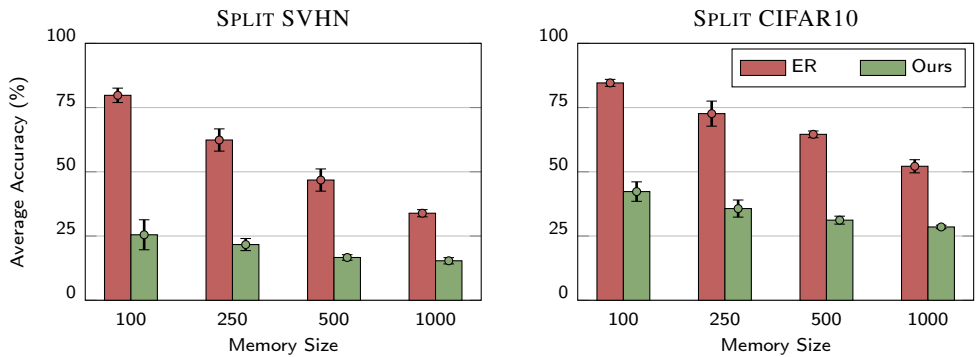
\begin{figure}
    \centering
    \begin{minipage}{.49\textwidth}
        \begin{tikzpicture}[declare function={barWidth=10pt; barShift=barWidth/2;}]
            \pgfplotsset{
                xtick=data, major x tick style=transparent, ticklabel style={font=\scriptsize\sffamily\sansmath, yshift=3pt},
                every axis label/.append style={font=\sffamily\sansmath\scriptsize},
                legend style={font=\scriptsize\sffamily, legend columns=-1, /tikz/every even column/.append style={column sep=0.5cm}},
            }
            \begin{axis}[
                    height=5cm, width=\linewidth,
                    ybar=0pt, bar width=barWidth,
                    xmin=100, xmax=1000, symbolic x coords={100, 250, 500, 1000}, enlarge x limits=0.1, xtick=data,
                    ymin=0, ymax=100,
                    ymajorgrids=true, ytick distance=25,
                    legend image code/.code={\draw[#1, draw=black] (0cm,-0.1cm) rectangle (0.6cm,0.1cm);},
                    title=\textsc{Split SVHN},
                    ylabel=Average Accuracy (\%),
                    xlabel=Memory Size, xlabel style={yshift=2pt},
                    enlarge x limits=0.15,
                ]
                \addplot+[mark=*, mark options={xshift=-barShift, mark size=1.5pt}, draw=black, fill=knred!80,
                    error bars/.cd, y dir=both, y explicit, error bar style={line width=1pt,solid, black},] table[x=memory, y=svhnerm, y error=svhners] {\memoryforgetting};
                \addplot+[mark=*, mark options={xshift=barShift, mark size=1.5pt}, draw=black, fill=kngreen!80,
                    error bars/.cd, y dir=both, y explicit, error bar style={line width=1pt,solid, black},] table[x=memory, y=svhngclm, y error=svhngcls] {\memoryforgetting};
            \end{axis}
        \end{tikzpicture}
    \end{minipage}\hspace{0.1cm}
    \begin{minipage}{.49\textwidth}
        \begin{tikzpicture}[declare function={barWidth=10pt; barShift=barWidth/2;}]
            \pgfplotsset{
                xtick=data, major x tick style=transparent, ticklabel style={font=\scriptsize\sffamily\sansmath, yshift=3pt},
                every axis label/.append style={font=\sffamily\sansmath\scriptsize},
                legend style={font=\scriptsize\sffamily, legend columns=-1, /tikz/every even column/.append style={column sep=0.5cm}},
            }
            \begin{axis}[
                    height=5cm, width=\linewidth,
                    ybar=0pt, bar width=barWidth,
                    xmin=100, xmax=1000, symbolic x coords={100, 250, 500, 1000}, enlarge x limits=0.1, xtick=data,
                    ymin=0, ymax=100,
                    ymajorgrids=true, ytick distance=25,
                    legend image code/.code={\draw[#1, draw=black] (0cm,-0.1cm) rectangle (0.6cm,0.1cm);},
                    title=\textsc{Split CIFAR10},
                    xlabel=Memory Size, xlabel style={yshift=2pt},
                    enlarge x limits=0.15,
                ]
                \addplot+[mark=*, mark options={xshift=-barShift, mark size=1.5pt}, draw=black, fill=knred!80,
                    error bars/.cd, y dir=both, y explicit, error bar style={line width=1pt,solid, black},] table[x=memory, y=cifarerm, y error=cifarers] {\memoryforgetting};
                \addplot+[mark=*, mark options={xshift=barShift, mark size=1.5pt}, draw=black, fill=kngreen!80,
                    error bars/.cd, y dir=both, y explicit, error bar style={line width=1pt,solid, black},] table[x=memory, y=cifargclm, y error=cifargcls] {\memoryforgetting};
                \legend{ER, Ours}
            \end{axis}
        \end{tikzpicture}
    \end{minipage}
    \caption{\small Average forgetting as a function of memory size on \textsc{Split SVHN} and \textsc{Split CIFAR10}.}
\end{figure}

\smallskip
\section{Hyper-parameters} \label{hyperparameters} Following \cite{lopez2017gradient}, we report the hyper-parameter grids considered in our experiments. These hyper-parameters are selected independently for each model, and the best values are given in parenthesis.

\begin{itemize}
	\item Finetune
	      \begin{itemize}
		      \item \texttt{optimizer: [Adam (Split SVHN, Split CIFAR10), SGD (Permuted MNIST, Rotated MNIST)]}
		      \item \texttt{learning rate: [0.0002, 0.001 (Split SVHN, Split CIFAR10,  Split CIFAR100, Split MiniImagenet), 0.01, 0.1 (Permuted MNIST, Rotated MNIST), 0.3, 1.0]}
	      \end{itemize}

	\item EWC \citep{kirkpatrick2017overcoming}
	      \begin{itemize}
		      \item \texttt{optimizer: [Adam, SGD (Permuted MNIST, Rotated MNIST, Split SVHN, Split CIFAR10)]}
		      \item \texttt{learning rate: [0.0002 (Split SVHN, Split CIFAR10), 0.001, 0.01, 0.1 (Permuted MNIST, Rotated MNIST), 0.3, 1.0]}
		      \item \texttt{regularization: [0.1, 1, 10 (Permuted MNIST, Rotated MNIST), 100 (Split SVHN, Split CIFAR10,  Split CIFAR100, Split MiniImagenet), 1000]}
	      \end{itemize}

	\item GEM \citep{lopez2017gradient}
	      \begin{itemize}
		      \item \texttt{optimizer: [Adam (Split SVHN, Split CIFAR10), SGD (Permuted MNIST, Rotated MNIST)]}
		      \item \texttt{learning rate: [0.0002, 0.001 (Split CIFAR10,  Split CIFAR100, Split MiniImagenet), 0.01, 0.1 (Permuted MNIST, Rotated MNIST), 0.3, 1.0]}
		      \item \texttt{margin: [0.0, 0.1, 0.5 (Permuted MNIST, Rotated MNIST), 1.0 (Split SVHN, Split CIFAR10,  Split CIFAR100, Split MiniImagenet)]}
	      \end{itemize}

	\item MER \citep{riemer2018learning}
	      \begin{itemize}
		      \item \texttt{optimizer: [SGD (Permuted MNIST, Rotated MNIST)]}
		      \item \texttt{learning rate: [0.0002, 0.001, 0.01, 0.1 (Permuted MNIST, Rotated MNIST), 0.3, 1.0]}
		      \item \texttt{within batch meta-learning rate (beta): [0.01 (Permuted MNIST, Rotated MNIST, Split CIFAR10,  Split CIFAR100, Split MiniImagenet), 0.03, 0.1, 0.3, 1]}
	      \end{itemize}

	\item ER \citep{chaudhry2019continual}
	      \begin{itemize}
		      \item \texttt{optimizer: [Adam (Split SVHN, Split CIFAR10,  Split CIFAR100, Split MiniImagenet), SGD (Permuted MNIST, Rotated MNIST)]}
		      \item \texttt{learning rate: [0.0002, 0.001 (Split SVHN, Split CIFAR10,  Split CIFAR100, Split MiniImagenet), 0.01, 0.1 (Permuted MNIST, Rotated MNIST), 0.3, 1.0]}
	      \end{itemize}

	\item GCL
	      \begin{itemize}
		      \item \texttt{optimizer: [Adam (Permuted MNIST, Rotated MNIST, Split SVHN, Split CIFAR10,  Split CIFAR100, Split MiniImagenet), SGD]}
		      \item \texttt{learning rate: [0.0002, 0.001 (Permuted MNIST, Rotated MNIST, Split SVHN, Split CIFAR10,  Split CIFAR100, Split MiniImagenet), 0.01, 0.1, 0.3, 1.0]}
		      \item \texttt{graph regularization: [0, 10, 50 (Split SVHN, Split CIFAR10,  Split CIFAR100, Split MiniImagenet), 100, 1000, 5000 (Rotated MNIST)]}
		      \item \texttt{context temperature: [0.1 (Permuted MNIST, Rotated MNIST), 0.3, 1 (Split SVHN, Split CIFAR10,  Split CIFAR100, Split MiniImagenet), 5, 10]}
		      \item \texttt{target temperature: [0.1, 0.3, 1, 5 (Permuted MNIST, Rotated MNIST, Split SVHN, Split CIFAR10, Split CIFAR100, Split MiniImagenet), 10]}
	      \end{itemize}
\end{itemize}

\end{document}